%% file: bare_jrnl_new_sample4.tex
\documentclass[lettersize,journal]{IEEEtran}
\usepackage{amsmath,amsfonts}
\usepackage{algorithmic}
\usepackage{array}
\usepackage[caption=false,font=normalsize,labelfont=sf,textfont=sf]{subfig}
\usepackage{textcomp}
\usepackage{stfloats}
\usepackage{url}
\usepackage{verbatim}
\usepackage{graphicx}
\usepackage{algorithm}
\usepackage{multicol}
\usepackage{multirow}
\usepackage{colortbl}
\usepackage{tcolorbox}
\usepackage{bm} 
\usepackage{hyperref}
\usepackage{booktabs}
\begin{document}

\title{Enhancing HOI Detection with Contextual Cues from Large Vision-Language Models}

\author{Yu-Wei Zhan,
        Fan Liu,~\IEEEmembership{Member,~IEEE},
        Xin Luo,
        Xin-Shun Xu,~\IEEEmembership{Senior Member,~IEEE},
        Liqiang Nie,~\IEEEmembership{Senior Member,~IEEE},
        Mohan Kankanhalli,~\IEEEmembership{Fellow,~IEEE}
\IEEEcompsocitemizethanks{
\IEEEcompsocthanksitem Yu-Wei Zhan is with Tsinghua University, China. Email: zhanyuweilif@gmail.com.
\IEEEcompsocthanksitem This work was done when Yu-Wei Zhan was a research intern at the National University of Singapore. 
\IEEEcompsocthanksitem Fan Liu and Mohan Kankanhalli are with National University of Singapore, Singapore. E-mail: liufancs@gmail.com, mohan@comp.nus.edu.sg.
\IEEEcompsocthanksitem Xin Luo and Xin-Shun Xu are with Shandong University, China. Email: luoxin.lxin@gmail.com, xuxinshun@sdu.edu.cn.
\IEEEcompsocthanksitem Liqiang Nie is with Harbin Institute of Technology (Shenzhen), China. E-mail: nieliqiang@gmail.com.
}
}


\markboth{Journal of \LaTeX\ Class Files,~Vol.~14, No.~8, August~2021}%
{Shell \MakeLowercase{\textit{et al.}}: A Sample Article Using IEEEtran.cls for IEEE Journals}

\maketitle
\begin{abstract}
Human-Object Interaction (HOI) detection aims at detecting human-object pairs and predicting their interactions. However, conventional HOI detection methods often struggle to fully capture the contextual information needed to accurately identify these interactions. While large Vision-Language Models (VLMs) show promise in tasks involving human interactions, they are not tailored for HOI detection. The complexity of human behavior and the diverse contexts in which these interactions occur make it further challenging. Contextual cues, such as the participants involved, body language, and the surrounding environment, play crucial roles in predicting these interactions, especially those that are unseen or ambiguous. Moreover, large VLMs are trained on vast image and text data, enabling them to generate contextual cues that help in understanding real-world contexts, object relationships, and typical interactions.
Building on this, in this paper we introduce ConCue, a novel approach for improving visual feature extraction in HOI detection. Specifically, we first design specialized prompts to utilize large VLMs to generate contextual cues within an image. To fully leverage these cues, we develop a transformer-based feature extraction module with a multi-tower architecture that integrates contextual cues into both instance and interaction detectors. Extensive experiments and analyses demonstrate the effectiveness of using these contextual cues for HOI detection. The experimental results show that integrating ConCue with existing state-of-the-art methods significantly enhances their performance on two widely used datasets.
The code is available at \href{https://github.com/yw-zhan/ConCue}{https://github.com/yw-zhan/ConCue}.
\end{abstract}


\input{sec/1_Introduction}

\input{sec/2_Related_work}

\input{sec/4_Method_v2}
\input{sec/5_Experiments}

\input{sec/6_Conclusion}

\bibliographystyle{IEEEtran}
\bibliography{sample-base}

\vspace{0em}

\end{document}

%% file: sec/1_Introduction.tex
\section{Introduction}
\label{sec:intro}
\IEEEPARstart{H}{uman}-Object Interaction (HOI) detection aims to accurately recognize humans and objects within images and subsequently understand the interactions between them. 
It holds immense potential for various downstream applications, significantly enhancing perception and comprehension capabilities within automated systems. Notable applications span diverse domains, including image and video captioning~\cite{chen2020say, you2016image}, visual surveillance~\cite{10454595}, and autonomous driving~\cite{Huang_2023_ICCV}.

\begin{figure}[t]
\centering
\begin{minipage}{1.0\linewidth}\centering
\centerline{\includegraphics[height=9.8cm]{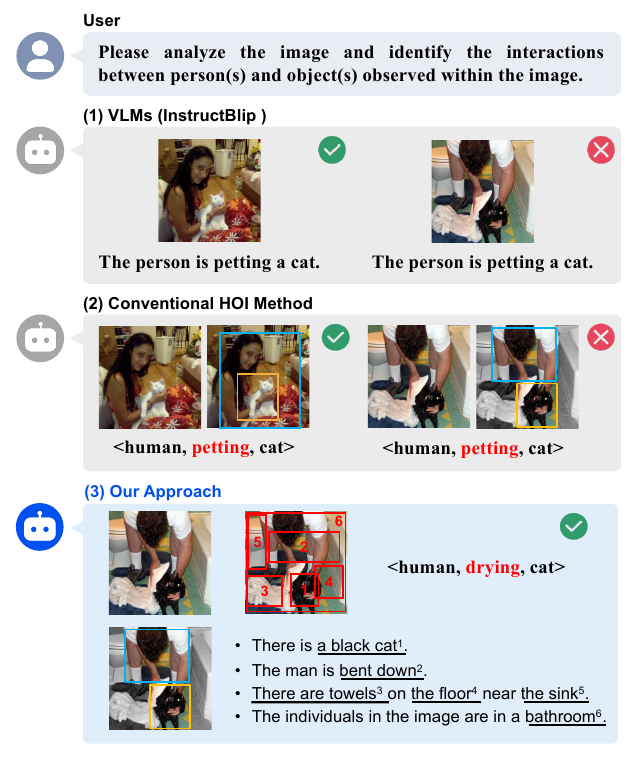}}
\end{minipage}
\vspace{0.0cm} 
\caption{An example illustrating the performance of HOI detection methods. (1) Large VLMs (e.g. InstructBlip) are not specifically designed for HOI detection, and (2) conventional HOI methods focus on visual information but overlook contextual information, leading to incorrect interaction classification. (3) Our proposed approach leverages contextual cues to achieve more accurate results.
}\label{intro}\medskip\end{figure}

Early HOI detection methods first identify and localize humans and objects within images. The region proposals associated with the detected humans and objects are then utilized to determine the type of interactions occurring between them~\cite{hou2020visual, HouY0PT21, li2020hoi, zhang2022efficient}. 
To simplify the training process, recent methods streamline the HOI detection task by formulating it as a set prediction problem~\footnote{Predicting a set of human-object interactions, where each element in the set includes a human, an object, and their interaction.}. By introducing queries, these methods detect HOI triplets in a one-stage manner~\cite{dong2022category, yuan2022detecting, TamuraOY21, iftekhar2022look}. To enhance the understanding and identification of interactions between humans and objects, recent research efforts have focused on leveraging relevant features and structures within images. For example, PViC~\cite{zhang2023exploring} introduces a cross-attention mechanism that selectively concentrates on the salient aspects of global features. This helps reduce the impact of noise from irrelevant regions, leading to more accurate HOI triplet representations. To further strengthen HOI predictions, some works take advantage of the inter-interaction semantic structure or the intra-interaction spatial structure between humans and objects~\cite{zhang2022exploring, ZhouQWSZ22}. 
More recently, recognizing the wealth of prior knowledge in pre-trained models, several methods have incorporated this prior language knowledge to enhance interaction comprehension~\cite {LiaoZLWL022, li2022improving}. Specifically, they employ handcrafted templates to convert each HOI triplet into a descriptive sentence, which is then fed into the CLIP text encoder to obtain embeddings for the HOI category descriptions, subsequently used as the classifier's weights.

Despite significant progress, conventional HOI detection methods still face challenges in fully comprehending the contextual nuances of interactions. 
Refer to Figure~\ref{intro} for an illustration. 
The two images depict ``a person is \emph{petting} a cat.'' and ``a person is \emph{drying} a cat.'', respectively. However, conventional methods identify the HOI triplet in both images as $<human, petting, cat>$. This is because both actions (petting and drying) look visually similar in images, as both involve a person touching or handling a cat. The spatial relationship and posture of the human and the cat could be nearly identical, making it difficult to differentiate between these two actions. Moreover, conventional methods typically rely on feature extraction techniques that may not capture the subtle differences between similar actions. In other words, the features extracted for petting and drying might overlap significantly, leading to misclassification. 
Large Vision-Language Models (VLMs) exhibit a nuanced comprehension of both visual and textual information~\cite{zhu2023minigpt, liu2023visual}. They can learn the correlations between visual and semantic elements from both images and corresponding textual descriptions.
While they are proficient in aligning visual and textual data, these models are not specifically designed to identify complex interactions across various scenarios. As illustrated in Figure~\ref{intro}, InstructBlip identifies both interactions as "a person is \emph{petting} a cat.". This bottleneck arises partly because HOI detection requires a fine-grained understanding of the specific interactions between humans and objects, which may not be sufficiently represented in the training data of VLMs. 

Contextual information plays a crucial role in enhancing visual understanding by providing additional clues and background details~\cite{WangZ23}. Large VLMs are trained on vast amounts of image and text data. They can generate contextual cues that help in understanding a wide range of real-world contexts, object relationships, and typical interactions. Accordingly, these contextual cues can improve HOI detection by offering additional layers of meaning that help disambiguate similar objects or interactions based on their surroundings. 
Based on the observation of ``there are towels'' in the image, especially within ``the bathroom'' scene, leads us to accurately conclude that the action is ``drying the cat'' rather than ``petting the cat.'' 
This is because, in the bathroom, a person using a towel on a cat is likely to dry the cat rather than pet it.
Leveraging these cues enables HOI detection models to gain a deeper understanding of human behavior and the underlying motivations in interactions, improving the accuracy of interaction inferences.
Moreover, by focusing on these contextual cues, models offer greater interpretability, elucidating the reasons behind specific interaction predictions, which is a step beyond mere recognition.

Inspired by the aforementioned considerations, we propose a novel approach called ConCue, which leverages the contextual cues derived from VLMs for HOI detection. Contextual cues, such as participants, body language, surrounding environment, and temporal features, play a crucial role in accurately identifying humans, objects, and complex interactions within an image. To generate these cues, we design a specialized set of prompts that derive the knowledge embedded in large VLMs.
To fully harness the power of these contextual cues, we introduce a cues-driven feature extraction method utilizing a multi-tower transformer architecture. This method is designed to guide and enhance the extraction of visual features by incorporating contextual cues, ultimately improving the accuracy of instance and interaction decoders.
Our proposed ConCue primarily focuses on feature extraction by integrating contextual cues for HOI detection. Thus, the developed context-aware instance and interaction decoders within ConCue can seamlessly replace the instance and interaction decoders in conventional HOI detection methods, thereby enhancing their performance. Additionally, by incorporating contextual cues into feature extraction, our approach can significantly improve the interpretability of existing methods in predicting complex human-object interactions.

The primary contributions of our paper can be summarized as follows:
\begin{itemize}
\item We highlight the limitations of conventional methods and large VLMs in HOI detection. To overcome these limitations, we propose a novel approach called ConCue, which utilizes contextual cues to enhance the performance of conventional methods.

\item We emphasize the critical role of contextual cues in identifying complex interactions. To generate these cues, we design a specialized set of prompts that derive the knowledge embedded in large
VLMs.

\item To fully harness the capabilities of large VLMs, we introduce a cue-driven visual feature extraction method that effectively guides and enhances feature extraction using contextual cues.

\item Our proposed ConCue has been validated on two widely recognized HOI detection benchmarks, demonstrating its effectiveness, interoperability, and the practical utility of contextual cues derived from large VLMs.
\end{itemize}

%% file: sec/2_Related_work.tex
\section{Related Work}

\subsection{Human-Object Interaction Detection}

Human-Object Interaction (HOI) detection involves recognizing humans/objects and classifying interactions between them within an image~\cite{LiuWGRLRLY22, LiLWLQXXFL23}. Conventional HOI detection methods fall into two categories: two-stage methods, which entail separate recognition and classification, and one-stage methods, which handle both tasks simultaneously.

\textbf{Two-stage methods}~\cite{WangYDWYT21, HouY0PT21, WangYYT20, ZhongDQT21, abs-2304-08114} first detect humans and objects in an image using independent detectors. Then, a well-designed classification module is employed to associate humans and objects and infer the HOI class for each human-object pair. These methods primarily focus on enhancing classification performance by leveraging additional information, including spatial features~\cite{UlutanIM20, LiLLWLLL20}, pose and body features \cite{WanZLLH19, GuptaSH19a, liu2022highlighting}, and semantic features \cite{BansalRSC20, LiuYC20, LiXLHXWFMCL20}. However, the computational efficiency of two-stage methods is often compromised due to their multi-step process.

\textbf{One-stage methods} streamline the process by combining human/object detection and interaction prediction within a single stage. Early one-stage methods \cite{WangYDK0S20} sought to extract relevant information from predefined interaction regions and accurately match the corresponding subjects and objects. For example, PPDM \cite{LiaoLWCQF20} adopts the keypoints to represent the HOI instance, while UnionDet \cite{KimCKK20} uses joint regions as predefined anchors to associate the humans and objects. Although these methods improve efficiency, they often require intricate post-processing for accurate matching.
Recently, transformer-based detectors, like DETR \cite{CarionMSUKZ20}, have shown remarkable performance in HOI detection. These methods~\cite{kim2023relational, zhang2022exploring, dong2022category, yuan2022detecting, kim2022mstr, fang2021dirv, MaWWW24} treat HOI detection as a set prediction task, employ query embeddings to represent HOI instances. Notably, QPIC~\cite{TamuraOY21} uses a single query 
embedding to capture only one HOI instance to avoid confusion between multiple instances. Furthermore, GEN-VLKT~\cite{LiaoZLWL022} uses two decoders to separately perform human/object detection and interaction prediction. This approach makes these two tasks independent of each other, leading to better performance. Additionally, CQL~\cite{xie2023category} introduces query designs associated with interaction categories, converting them into image-specific category representations, and learns through an auxiliary image-level classification task.

\subsection{Pretrained Vision-Language Models}
In recent years, the pre-trained Vision-Language Models (VLM) have demonstrated significant transferability to various downstream tasks~\cite{LuddeckeE22,yuan2022rlip,YangDALFGYL23,WanLZT023}. Due to their powerful capabilities, such as the CLIP model, they are employed to improve HOI detection.
Key approaches include: (1) leveraging VLM's powerful representation capabilities by aligning features or logits with the embedded image features in VLM~\cite{LiaoZLWL022}; (2) utilizing VLM's textual prior knowledge by using the text labels embedded in VLM to initialize classifiers \cite{yuan2022rlip, WanLZT023}.

Most recently, there has been a surge of interest in large VLMs~\cite{zhu2023minigpt, liu2023visual}. These models have achieved remarkable breakthroughs in image and language understanding through extensive training data and advanced architectures. They can accurately generate descriptions of images, deeply understand the semantics of text, and effectively connect images to natural language.
Large VLMs have broad prospects and potential in HOI detection. UniHOI \cite{CaoTSCYLX23} utilizes the representational capabilities of VLMs to extract enhanced feature vectors of images. These richer and higher-level feature representations enable the model to better classify interactions between humans and objects. While this strategy excels in feature representation, it still has limitations in complex semantic reasoning, falling short in fully understanding the intricate relationships and interaction logic between objects. 
In this paper, we harness the generative capabilities of large VLMs and use prompts to guide them in generating contextual cues that capture the context of interactions within the image. We then integrate these visual cues with general image features to enhance interaction understanding.

%% file: sec/4_Method_v2.tex

\section{Preliminaries}
In traditional Transformer-based architectures for HOI detection, the process typically involves \textbf{image encoder}, \textbf{instance detector}, and \textbf{interaction detector}. 

Initially, the input image is processed through a CNN to generate the feature map $X\in\mathbb{R}^{C \times H \times W}$, where $C$ represents the number of channels, and $H$ and $W$ represent the height and width, respectively. The feature map is then divided into fixed-size patches and flattened into vectors using the mapping function. These vectors $\{X_1, X_2,..., X_n\}$ are arranged into a sequence based on their spatial positions, where $n$ is the number of patches. Finally, the feature representation of each patch is combined with its corresponding positional encoding $\zeta_{pos} \in\mathbb{R}^{n \times d}$. Through these steps, the input image is transformed into a sequential representation. Utilizing multiple Transformer encoder layers, the visual features of the image are obtained:
\begin{equation}\label{xv}
{X}_I = \mathcal{F}_{\theta_I}(\ [X_1; X_2; ...; X_n] + \zeta_{pos}), 
\end{equation}
where $\theta_I$ is the parameter of the image encoder.

For instance detection, the learnable human queries $Q_h\in\mathbb{R}^{N_q \times D}$ and object queries $Q_o\in\mathbb{R}^{N_q \times D}$ are introduced to represent humans and objects, respectively. Here, $N_q$ denotes the number of query vectors and $D$ is the dimension of queries. As each query captures at most one human-object pair, $N_q$ is set to be sufficiently large to ensure it always exceeds the actual number of human-object pairs within the image. Position embeddings $Q_p$ are utilized to match individuals and objects in pairs. The \textbf{instance detector} can be described as follows:
\begin{equation}\label{instance}
[{E}_h, {E}_o] = \mathcal{F}_{\theta_H}({X}_I, Q_h+Q_p, Q_o+Q_p), 
\end{equation}
where $\theta_H$ is the parameter of the instance detector. 
$E_{h}$ and $E_o$ represent tokens highly correlated with the position information of each pair. These outputs are then utilized to generate $N_q$ predictions, including bounding box pairs for person-object ($B_h \in \mathbb{R}^{N_q \times 4}$, $B_o \in \mathbb{R}^{N_q \times 4}$) and object class scores $C_o \in \mathbb{R}^{N_q \times l_o}$, where $l_o$ is the number of object classes. Note that two-stage methods typically employ off-the-shelf encoders and instance detectors, which are fine-tuned and then frozen. These methods separate the processes of object detection and interaction classification into two distinct steps. In contrast, one-stage methods often perform object detection and interaction classification concurrently.


For more challenging interaction recognition tasks, many methods utilize an additional detector network to perform classification predictions using $E_o$ and $E_h$. Similarly, the \textbf{interaction detector} can be described as follows:
\begin{equation}\label{interaction}
{E}_{inter} = \mathcal{F}_{\theta_R}({X}_I, {E}_h, {E}_o), 
\end{equation}
where $\theta_R$ represents the parameters of the interaction detector. 
$E_{inter}$ is the output of the interaction detector, used for classification.


\begin{figure*}[t]
\centering
\begin{minipage}{1.0\linewidth}\centering
\centerline{\includegraphics[height=7.6cm]{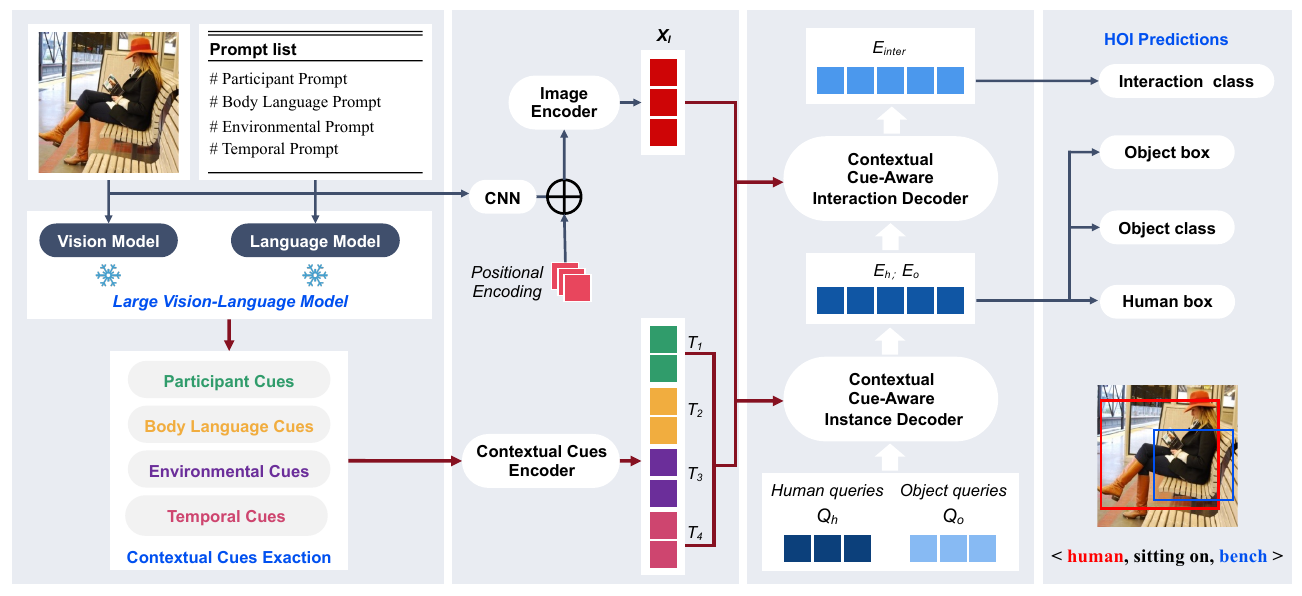}}
\end{minipage}
\vspace{0.0cm} 
\caption{{The overall framework of ConCue.} We first utilize a large VLM (e.g., InstructBlip) to generate contextual cues, such as participant cues, body language cues, environmental cues, and temporal cues within images. These cues are then incorporated into the contextual cue-aware instance and interaction decoders to enhance feature extraction for HOI detection.}\vspace{-0.0cm} \label{framework}\medskip\end{figure*}

\section{ConCue}
\subsection{Overview}
Conventional HOI detection methods, such as transformer-based approaches, excel at extracting features from an image by leveraging the attention mechanism, which captures global contextual relationships among image patches. While effective, these methods typically focus solely on visual information like color, shape, and spatial relations. However, contextual information provides essential background and supplementary knowledge that enhances the model's ability to accurately detect and interpret human-object interactions. By considering the broader environment, typical object uses, and expected human behaviors, context helps HOI detection models make more informed and precise predictions, ultimately improving their performance and robustness.

In this paper, we introduce the use of contextual cues to enhance the performance of conventional HOI detection methods. While visual features provide essential details about what is present in the image, contextual cues derived from large VLMs offer a deeper, more informed perspective by incorporating semantic knowledge, guiding attention, and disambiguating visual information. 
The framework of our proposed ConCue is illustrated in Figure~\ref{framework}. We generate contextual cues by using a specialized set of prompts applied to VLMs, which provide various clues and insights into the context surrounding interactions. To incorporate these contextual cues into HOI detection, we develop a transformer-based feature extraction module that is utilized within both the contextual cue-driven instance and interaction decoders. This module can guide the extraction of visual features and enhance them with contextual insights.
As a result, these decoders can more accurately detect humans, objects, and their interactions by fully leveraging the power of contextual cues.
Detailed explanations of these components are elaborated upon in the subsequent sections.


\subsection{Contextual Cues Generation}

Given the complexity and variability of real-world scenarios, it is essential to incorporate contextual cues to enrich the information available for HOI detection. Contextual cues, including participant cues, body language cues, environmental cues, and temporal cues, provide a comprehensive understanding of the interaction context, helping to accurately interpret the intricate dynamics between humans and objects in various scenarios~\footnote{These four types of cues are predefined. Incorporating additional types of cues could potentially further enhance the performance of our ConCue by providing even richer contextual information.}. Next, we will introduce how to generate contextual cues from the large VLM.

\begin{figure*}[t]
\centering
\begin{minipage}{1.0\linewidth}\centering
\centerline{\includegraphics[height=6cm]{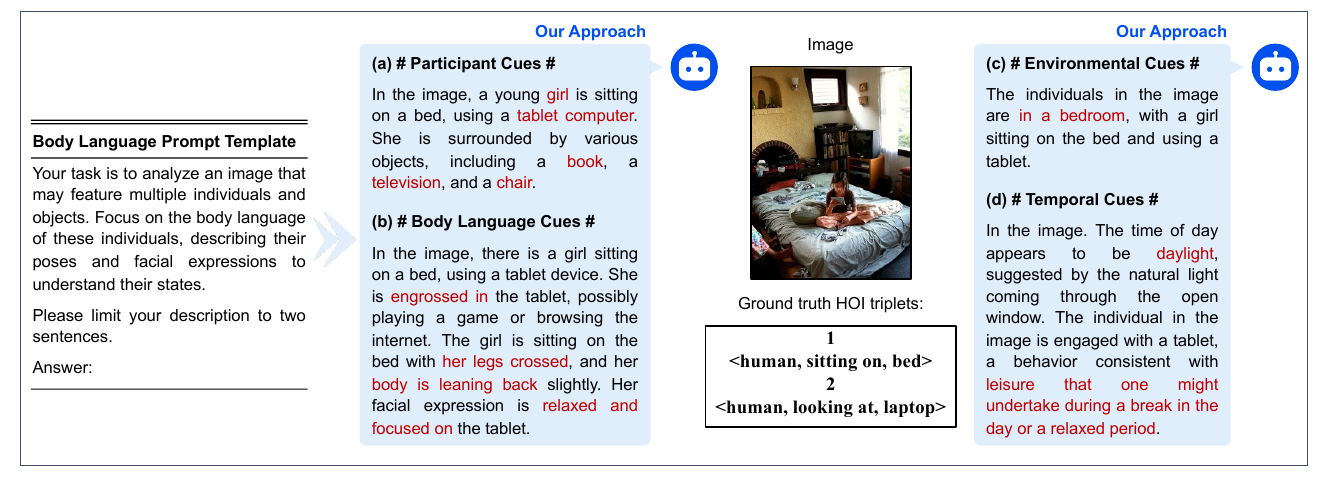}}
\end{minipage}
\vspace{0.0cm} 
\caption{\textbf{An illustration of generating contextual cues}, including (a) \# Participant Cues \#, (b) \# Body Language Cues \#, (c) \# Environmental Cues \#, and (d) \# Temporal Cues \#.}\label{cues}\medskip\end{figure*}

\subsubsection{Prompts with Large VLM}
To generate contextual cues for an image using the large VLM, we propose a specialized set of prompts. This set includes the participant prompt, body language prompt, environmental prompt, and temporal prompt. We input the prompt along with the image into the large VLM, which then generates the contextual cues in text form. By inputting these prompts along with the image into the large VLM, the model generates the contextual cues in text form. As illustrated in Figure~\ref{cues}, we provide an example of a body language prompt to demonstrate its details.

\subsubsection{Contextual Cues}
Each type of cue captures a distinct aspect of the interaction, contributing to a more accurate and robust detection model. The details of the generated cues for the four types are shown in Figure~\ref{cues}.
\begin{itemize}

\item \textbf{Participant Cues.} These cues focus on the humans and objects with which they potentially interact within the image. As illustrated in Figure~\ref{cues}(a), they guide the large VLM to identify and emphasize the key humans and relevant objects in the scene. 

\item \textbf{Body Language Cues.} These refer to the posture and facial expressions of individuals. As shown in Figure~\ref{cues}(b), posture can reveal a person's current state and intentions, while facial expressions are key indicators of emotional states. These cues are invaluable for understanding the range of emotional responses to interactions across different contexts.

\item \textbf{Environmental Cues.}  As illustrated by Figure~\ref{cues}(c), these refer to the contextual setting surrounding individuals and objects. The environment in which a person is situated plays a crucial role in HOI detection, providing essential context and background for the interactions taking place.

\item \textbf{Temporal Cues:} From Figure~\ref{cues}(d), it can be seen that temporal cues provide information about the time of day or historical period, significantly influencing prediction. The position of shadows, style of clothing, and type of activity can indicate a specific time or era, shaping our understanding of the interaction. 
\end{itemize}

\subsection{Contextual Cue-Enhanced Feature Extraction} \label{CDCA}
To integrate contextual cues derived from large VLMs into conventional HOI detection methods, we develop a transformer-based feature extraction module with a multi-tower architecture. This module primarily consists of (1)~\textbf{multiple cue-driven decoders} that extract cue-aware visual features from various contextual cues and (2) a \textbf{feature fusion component} that combines these cue-aware features with the original visual features. It is utilized within both the context-aware instance and interaction decoders of ConCue to guide and enhance the extraction of visual features. 

\begin{figure}[t]
\centering
\begin{minipage}{1.0\linewidth}\centering
\centerline{\includegraphics[height=7.3cm]{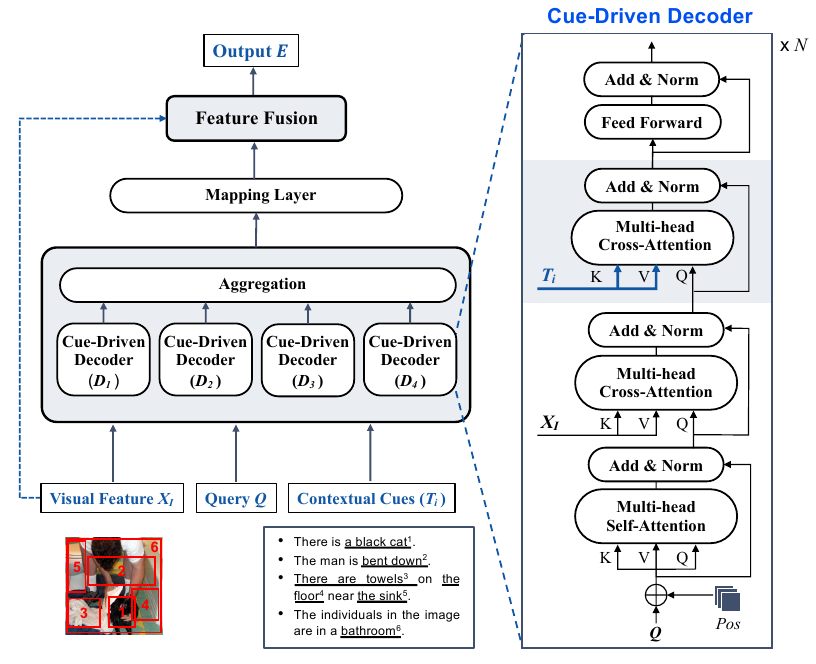}}
\end{minipage}
\vspace{-0.0cm} 
\caption{Structure of the transformer-based contextual cue-enhanced feature extraction module. This module primarily consisted of multiple cue-driven decoders and a feature fusion component.}\vspace{-0.0cm} \label{decoder}\medskip\end{figure}
\textbf{Multiple Cue-Driven Decoders.}
As illustrated in Figure~\ref{decoder}, the module features a multi-tower structure, with each tower associated with a different cue-driven decoder.
Each cue-driven decoder is responsible for extracting visual features corresponding to a specific cue feature $T_i \in \mathcal{T}$. Taking the cue feature $T_i$ as an example, we feed the queries $Q$ into the module, passing them through the first two attention blocks of one tower:
\begin{equation}
{Q}' = {\emph{Cross\text{-}Attn}}({X}_I, \ {\emph{ Self\text{-}Attn}}({Q})),
\label{loss1}
\end{equation}
where ${X}_I$ is the visual feature of an image. 
Next, we introduce a new Cross-Attention layer to the vanilla transformer to obtain the cue-aware visual feature. Specifically, we replace the keys (K) and values (V) with $T_i$:
\begin{equation}
\begin{split}
Q'' & = {\emph{ Cross\text{-}Attn}}(T_i, \ Q')\\
& = softmax(\frac{{Q}'\  {T}_{i}^T}{\sqrt{d_t}}{T}_{i}). 
\end{split}\label{loss2}
\end{equation}
Here, $d_t$ is the dimension of $T_i$. The output $Q''$ is then fed into a feed-forward network:
\begin{equation}
{E}_i  = {\emph{ FFN}}(Q'').
\label{loss21}
\end{equation}

Finally, we aggregate the output of multiple cue-driven decoders using the $\emph{Agg}(\cdot)$ function, which performs aggregation through either average pooling or concatenation, based on the specific requirements of instance and interaction decoders. Specifically, we aggregate cue-aware features using average pooling in the instance decoder and concatenation in the interaction decoder. The aggregation process is detailed below:
\begin{equation}
\begin{split}
{\hat{E}} = \emph{Agg}(E_1, E_2, \dots, E_n).
\end{split}
\label{loss3}
\end{equation}

\textbf{Feature Fusion Component.} 
The original visual features contain detailed information directly derived from the image, such as textures, colors, and spatial relationships, which are essential for accurate object and interaction recognition. Cue-aware features, on the other hand, are enriched with contextual insights that provide a broader understanding of the scene. By fusing these two types of features, the model benefits from both detailed visual information and higher-level contextual understanding, resulting in a more comprehensive representation.
Specifically, we first reshape the visual features ${X}_I$ into $N_q \times D$ dimensions and then add them to $\hat{E}$. The final feature representations, $E$, are obtained through a mapping layer to project the combined features into a unified representation, denoted as $Proj(\cdot)$:
\begin{equation}
\begin{split}
{E}  = &Proj({X}_I + \hat{E}),
\end{split}
\label{loss4}
\end{equation}
This mapping layer ensures that the features are optimally integrated for downstream tasks.

\subsection{HOI Encoders and Decoders} 

\subsubsection{Contextual Cues and Image Encoders}
To encode the four types of contextual cues, we specifically designed a contextual cues encoder comprising a stack of Transformer encoder layers. To efficiently capture intricate semantic nuances and contextual dependencies within textual data and expedite model convergence, we initialized the parameters using RoBERTa, which is pre-trained on extensive text corpora, and subsequently froze them. Additionally, to adapt the RoBERTa model to our specific task, we incorporated three additional learnable encoder layers after the frozen RoBERTa layers. The features extracted from different contextual cues are represented as $\{T_i\}_{i=1}^n$, where $n$ is the number of cue types.

Our proposed method aims to enhance visual understanding by leveraging contextual cues for HOI detection. To this end, we integrated our method into various base models~ \cite{TamuraOY21, LiaoZLWL022, KimJC23, ning2023hoiclip} and employed the same image encoder used in each base model. 



\subsubsection{Contextual Cue-Aware Instance Decoder} \label{detector}
The purpose of the instance decoder is to decode the image feature into $N_q$ predictions. To obtain the contextual cue-enhanced output $E_h$ and $E_o$, we input
$Q_h$, $Q_o$, and contextual cues into the contextual cue-enhanced feature extraction module, as described in Section~\ref{CDCA}. 
Consequently, Eq. \ref{instance} is revised as follows: 
\begin{equation}
[{E}_h, {E}_o] = \mathcal{F}_{\theta_H^{'}} (X_I, Q_h+Q_p, Q_o+Q_p, \mathcal{T}), 
\end{equation}
By leveraging contextual cue features, our instance decoder can capture rich contextual information, resulting in more robust instance detection.

\subsubsection{Contextual Cue-Aware Interaction Decoder}
The interaction detector is responsible for predicting interaction classes for human-object pairs. Specifically, we initialize a set of interaction queries $Q_{inter} \in \mathbb{R}^{N_q \times D}$. Similar to the instance decoder, these queries are then fed into the contextual cue-enhanced feature extraction module. The output of the interaction decoder is denoted as $E_{inter}$. The description of the interaction decoder in Eq. \ref{interaction} is updated as follows:
\begin{equation}
{E}_{inter} = \mathcal{F}_{\theta_R^{'}}({X}_I, {E}_h, {E}_o, \mathcal{T}), 
\end{equation}

\textbf{Discussion}. One key distinction between our approach and previous methods~\cite{TamuraOY21, LiaoZLWL022} lies in the incorporation of contextual cues alongside visual features. Traditional HOI detection methods typically rely solely on visual features, which can sometimes limit their ability to fully understand the complexities of human-object interactions. In contrast, our contextual cue-aware decoders leverage a dual-modality approach, utilizing both visual features and contextual cue features.

The inclusion of contextual cues offers several advantages. These cues provide critical insights that are essential for accurately interpreting interactions, as they incorporate prior knowledge about the scene, objects, and potential interactions. By integrating these cues into the feature extraction process, our approach not only enhances the richness of the extracted visual features but also guides the model toward a more nuanced understanding of interactions, ultimately leading to more robust and accurate HOI detection.

\subsection{Model Training and Optimization}

As illustrated in Figure \ref{framework}, our method can be seamlessly integrated with any transformer-based HOI baseline approach. Integration is straightforward: during training, the contextual cue-aware decoders replace the original decoders in base models for HOI detection. The base model's original loss, $\mathcal{L}_{base}$, remains unchanged. Thus, the final training loss is then defined as:
\begin{equation}
{\mathcal{L}} = \mathcal{L}_{base}.
\end{equation}

In addition, our approach aims to leverage the knowledge embedded in the large VLMs to generate contextual cues, denoted as $\mathcal{T}$, which are incorporated into the training process. Therefore, our optimization objective is:
\begin{equation}\label{R-2}
\begin{split}
& \min \mathcal{L}(GT, \mathcal{F}_{\theta}({X}_I,  \mathcal{T}, {Q}_h, {Q}_o, {Q}_p)).\\
\end{split}
\end{equation}
where $\theta$ is the parameter of the whole framework and $GT$ is the ground truth.

%% file: sec/5_Experiments.tex
\section{Experiments}
To comprehensively assess the performance of ConCue, we conducted experiments on two real-world datasets to address several critical questions: Firstly, we compared ConCue's efficacy against large VLMs. Secondly, we evaluated its performance relative to conventional HOI methods. Additionally, we examined ConCue's effectiveness in zero-shot scenarios. Lastly, we scrutinized the efficacy of each component within ConCue and elucidated the reasons behind their effectiveness. 

In this section, we present the datasets and experimental setup before showcasing the results and providing an in-depth analysis of the aforementioned research inquiries.

\subsection{Experimental Setting}
\subsubsection{Datasets} We conducted experiments on two widely used public benchmarks: HICO-Det~\cite{chao2018learning} and V-COCO~\cite{gupta2015visual}. 
HICO-Det consists of $47,776$ images, with $38,118$ images designated for training and $9,658$ images for testing. This dataset includes annotations for $600$ HOI categories, which comprise $80$ object categories and $117$ actions. V-COCO serves as a subset of the COCO dataset. This subset comprises a total of $10,396$ images, split into $5,400$ images for training and $4,964$ images for testing. The annotations provided in V-COCO cover $263$ HOI categories, including $80$ object categories and $29$ actions.

\subsubsection{Evaluation Metrics}
To evaluate the effectiveness of our proposed ConCue, we compared its performance with \textbf{conventional HOI detection methods} and \textbf{InstructBlip}.

For comparison with conventional HOT detection methods, we employed the mean Average Precision (mAP) as our metric. Following the methodologies in~\cite{LiaoZLWL022, kim2023relational,dong2022category, kim2022mstr}, we determined the accuracy of an HOI triplet prediction using the following criteria: first, the Intersection over Union (IoU) values of the predicted the human and object bounding boxes must exceed $0.5$ \emph{w.r.t.} the Ground Truth box; second, the predicted HOI classes must be accurate. 

Empirically, large VLM cannot effectively perform ranking tasks that require ordering multiple predictions. For example,  LVLM, like InstructBlip, often generates only a single HOI triplet per image. This limitation necessitates using top-1 accuracy, which evaluates the correctness of only the first prediction, as the evaluation metric.


\begin{table*}[htp]
\centering \small
\caption{Comparisons of four baseline models with and without our ConCue on the HICO-Det and V-COCO datasets.}
\begin{tabular}{l>{\hfil}p{60pt}<{\hfil}>{\hfil}p{60pt}<{\hfil}>{\hfil}p{33pt}<{\hfil}>{\hfil}p{33pt}<{\hfil}>{\hfil}p{37pt}<{\hfil}cc} 
\toprule \multirow{2}{*}{Method} & \multirow{2}{*}{Pipeline} & \multirow{2}{*}{Backbone} & \multicolumn{3}{c}{ HICO-Det } & \multicolumn{2}{c}{ V-COCO } \\
\cmidrule(r){4-6}
& & & Full & Rare & Non-Rare & { $A P_{role}^{S1}$ } & { $A P_{role}^{S2}$ }\\
\midrule
QPIC (2021) \cite{TamuraOY21} & single-branch & ResNet-101 & 29.90 & 23.92 & 31.69 & 58.3 & 60.7 \\
\rowcolor{blue!8} {QPIC+ConCue} & single-branch & ResNet-101 & \textbf{32.99} & \textbf{27.65} & \textbf{34.58} & \textbf{61.1} &  \textbf{62.0} \\
GEN-VLKT (2022) \cite{LiaoZLWL022} & two-branch & ResNet-50 &  33.75 & 29.25 & 35.10& 62.4 & 64.5 \\
\rowcolor{blue!8} {GEN-VLKT+ConCue} & two-branch & ResNet-50 & \textbf{35.58} & \textbf{34.64} & \textbf{36.19} & \textbf{63.9} & \textbf{65.2} \\
MUREN (2023) \cite{KimJC23} & three-branch & ResNet-50  & 32.87 &28.67& 34.12& 68.8 &71.0 \\
\rowcolor{blue!8} {MUREN+ConCue} & three-branch & ResNet-50 &  \textbf{35.72} & \textbf{33.25} & \textbf{36.46} & \textbf{69.7 }&  \textbf{71.9} \\
HOICLIP (2023) \cite{ning2023hoiclip} & two-branch & ResNet-50 & 34.69 & 31.12& 35.74& 63.5& 64.8 \\
\rowcolor{blue!8} {HOICLIP+ConCue} & two-branch & ResNet-50 & \textbf{37.46} & \textbf{36.75} & \textbf{37.68} & \textbf{65.1} & \textbf{66.5} \\
\bottomrule
\end{tabular}
\vspace{0.0cm} 
\label{regular_performance_1}
\end{table*}
\subsubsection{Base Models}
In this work, we chose four representative methods as base models to validate the effectiveness of our approach.

\begin{itemize}
\item \textbf{QPIC}~\cite{TamuraOY21} is the first to introduce the transformers framework into the HOI task. It utilizes a CNN backbone to extract image features and employs a transformer decoder to detect humans and objects and to handle interaction classification. It uses the same transformer decoder for both instance detection and interaction detection.




\item \textbf{GEN-VLKT}~\cite{LiaoZLWL022} is a dual-branch approach utilizing two decoder branches. The first branch is tasked with detecting human-object pairs, while the second branch focuses on classifying the interactions between these pairs.


\item \textbf{MUREN}~\cite{KimJC23} introduces a novel method featuring a three-branch architecture, which separately focuses on human detection, object detection, and interaction classification. This method innovatively incorporates relational context information to enhance relation reasoning, marking a significant advancement in handling complex interactions within images.


\item \textbf{HOICLIP}~\cite{ning2023hoiclip} is a recent outstanding dual-branch work that involves the pre-trained vision-language model. It leverages the textual prior knowledge from CLIP, using text labels embedded in CLIP to initialize classifiers. This enhancement is particularly beneficial for handling unseen interaction classes. 

\end{itemize}

To verify its effectiveness, the proposed ConCue is integrated into the base models. Due to the varying number of decoder branches in the base models, the required decoder branches differ. Specifically, we employed 1, 2, 3, and 2 decoder branches for \textbf{QPIC+ConCue}, \textbf{GEN-VLKT+ConCue}, \textbf{MUREN+ConCue}, and \textbf{HOICLIP+ConCue}, respectively, to accommodate their respective architectures.

\subsubsection{Implementation Details}
In order to ensure a fair comparison, we adopt the same basic framework, parameters, and the same variant of CLIP used in the base model integrated with ConCue during the training process. We selected InstructBLIP \cite{dai2023instructblip} as the large VLM. During the whole training, we utilized the AdamW optimizer with an initial learning rate of $5e\mbox{-}5$. The batch size is $16$. All training processes are executed on $4$ NVIDIA A800 GPUs.

\subsection{Comparison with Base Models}

To evaluate the effectiveness of ConCue, we conducted comparative analyses by integrating our approach into four base models, specifically replacing their decoders with our proposed method under regular settings. Our evaluation utilized the following dataset configurations: for the HICO-Det dataset, we performed evaluations across three distinct category sets: (1) the complete set of $600$ HOI triplets, (2) the rare set comprising $138$ HOI categories with fewer than $10$ instances in the training data, and (3) the non-rare set, which includes the remaining $462$ HOI categories. For the V-COCO dataset, results are reported for two evaluation scenarios: S1, which includes $29$ action categories encompassing $4$ body movements, and S2, which consists of $25$ action categories without any no-object HOI categories. 

Table~\ref{regular_performance_1} summarizes the performance of four base HOI methods before and after integrating our approach. (1) QPIC, the first method to introduce the transformer framework into the HOI task, uses the same decoder for both instance detection and interaction detection. By incorporating the Cue-Driven Visual Feature Extraction Module, we achieved a significant performance improvement of 3.09 mAP.
(2) GEN-VLKT and HOICLIP are representative two-branch methods that employ separate decoder branches to detect human-object pairs and classify interactions. Our method resulted in performance gains of $1.83$ and $2.77$ mAP, respectively, establishing new state-of-the-art results on HICO-DET (see Table \ref{regular_performance_1}).
3) MUREN, the first method to employ three branches for person detection, object detection, and interaction classification. It achieved a $2.85$ mAP improvement when combined with our ConCue. 
On the V-COCO dataset, the performance improvements were similar, with increases of $2.8$, $1.6$, $0.9$, and $1.6$ mAP for QPIC, GEN-VLKT, MUREN, and HOICLIP on S1, respectively. However, the improvements on V-COCO were not significant compared to HICO-Det. This is likely due to the smaller amount of training data in V-COCO, which presents a challenge for effectively training transformer architectures.

Overall, the performance improvements across the four base models demonstrate the effectiveness of our proposed method. This is because our method can effectively leverage contextual cues to guide and enhance the extraction of visual features, thereby improving the performance of HOI detection.

\begin{table}[tp]
\centering \small
\caption{{Comparisons with state-of-the-art methods on the HICO-Det dataset.}}
\begin{tabular}{l>{\hfil}p{44pt}<{\hfil}>{\hfil}p{23pt}<{\hfil}>{\hfil}p{23pt}<{\hfil}>{\hfil}p{36pt}<{\hfil}} 
\toprule {Method} & {Backbone} & Full & Rare & Non-Rare \\
\midrule
VCL \cite{hou2020visual} & ResNet-50 & 23.63 & 17.21 & 25.55 \\
ATL \cite{HouY0PT21} & ResNet-50 & 23.67 & 17.64 & 25.47 \\
FCL \cite{hou2021detecting} & ResNet-50 & 24.68 & 20.03 & 26.07 \\
QPIC \cite{TamuraOY21} & ResNet-101 & 29.90 & 23.92 & 31.69 \\
CDN \cite{zhang2021mining} & ResNet-101 & 32.07 & 27.19 & 33.53 \\
SSRT \cite{iftekhar2022look} & ResNet-101 & 31.34 & 24.31 & 33.32  \\
IF \cite{liu2022interactiveness} & ResNet-50 & 33.51 & 30.30 & 34.46  \\
GEN-VLKT \cite{LiaoZLWL022} & ResNet-50 &  33.75 & 29.25 & 35.10 \\
Zhou $et \ al.$ \cite{zhou2023learning}&  ResNet-101 & 34.42& 30.03& 35.73 \\
OpenCat \cite{zheng2023open}& ResNet-101& 32.68& 28.42& 33.75 \\
ADA-CM \cite{lei2023efficient}&ResNet-50 &33.80 &31.72 &34.42  \\
PViC \cite{zhang2023exploring}& ResNet-50& 34.69& 32.14& 35.45 \\
MUREN & ResNet-50 & 32.87 &28.67& 34.12\\
RLIPv2 \cite{yuan2023rlipv2} & ResNet-50 & 33.29 & 27.27 & 35.08 \\
HOICLIP \cite{ning2023hoiclip} & ResNet-50 & 34.69 & 31.12& 35.74 \\
\hline 
\rowcolor{blue!8} {MUREN+ConCue} & ResNet-50 &  35.72 & 33.25 & 36.46  \\
\rowcolor{blue!8} HOICLIP+ConCue & ResNet-50 & \textbf{37.46} & \textbf{36.75} & \textbf{37.68} \\
\bottomrule
\end{tabular}
\vspace{0.0cm} 
\label{regular_performance}
\end{table}

\begin{table}
\centering \small
\caption{{Comparisons with state-of-the-art methods on the V-COCO dataset.}}
\begin{tabular}{l>{\hfil}p{50pt}<{\hfil}cc} 
\toprule{Method} & {Backbone} & { $A P_{role}^{S1}$ }&{ $A P_{role}^{S2}$ }\\
\midrule
VCL (2020) \cite{hou2020visual} & ResNet-50 & 48.3 & - \\
FCL (2021) \cite{hou2021detecting} & ResNet-50 &  52.4 & - \\
QPIC (2021) \cite{TamuraOY21} & ResNet-101 &  58.3 & 60.7 \\
CDN (2021) \cite{zhang2021mining} & ResNet-101 & 63.9 & 65.9 \\
SSRT (2022) \cite{iftekhar2022look} & ResNet-101 & 65.0 & 67.1 \\
IF (2022) \cite{liu2022interactiveness} & ResNet-50 & 63.0 & 65.2 \\
GEN-VLKT (2022) \cite{LiaoZLWL022} & ResNet-50 &  62.4 & 64.5 \\
Zhou $et \ al.$ (2023) \cite{zhou2023learning}&  ResNet-101 & 62.4& 68.1 \\
OpenCat (2023) \cite{zheng2023open}& ResNet-101& 61.9& 63.2 \\
ADA-CM (2023) \cite{lei2023efficient}&ResNet-50 &56.1&61.5 \\
PViC (2023) \cite{zhang2023exploring}& ResNet-50& 62.8& 67.8 \\
MUREN (2023) & ResNet-50 & 68.8 &71.0\\
RLIPv2 (2023) \cite{yuan2023rlipv2} & ResNet-50 & 63.8 &66.4 \\
HOICLIP (2023) \cite{ning2023hoiclip} & ResNet-50 & 63.5& 64.8 \\
\hline 
\rowcolor{blue!8} {MUREN+ConCue} & ResNet-50 & \textbf{69.7}&  \textbf{71.9} \\
\rowcolor{blue!8} HOICLIP+ConCue & ResNet-50 & 65.1 & 66.5 \\
\bottomrule
\end{tabular}
\vspace{0.0cm} 
\label{regular_performance_coco}
\end{table}

\subsection{Comparison with Regular HOI Detection}
To further evaluate the performance of ConCue, we conducted a comparison with other state-of-the-art methods under regular settings. The results on the HICO-Det and V-COCO datasets are presented in Tables \ref{regular_performance} and \ref{regular_performance_coco}, respectively, with MUREN and HOICLIP chosen as base models.

For the HICO-Det dataset, HOICLIP+ConCue significantly outperformes all other HOI methods. Specifically, MUREN, when enhanced with the ConCue, achieves a substantial performance improvement, bringing it on par with the state-of-the-art methods. Compared to the advanced PViC method, HOICLIP+ConCue achieves a gain of $8.0\%$ mAP in the FULL setting, with a margin of $2.77$ mAP. The results demonstrate the effectiveness of our method, emphasizing the value of leveraging contextual visual cues to identify interactions.  Particularly noteworthy is the performance in rare classes, where HOICLIP+ConCue achieves an mAP of $36.75$, significantly surpassing other methods, even in the FULL setting. This finding indicates that our method benefits from the knowledge acquired from LVLMs, improving the model's robustness. Even in scenarios with limited training samples, our method exhibits strong performance. For the V-COCO dataset, HOICLIP+ConCue achieved AP scores of $65.1$ in scenario S1 and $66.5$ in scenario S2, performing on par with previous state-of-the-art methods. MUREN+ConCue showed significant improvements with AP scores of $69.7$ in S1 and $71.9$ in S2, surpassing other methods.

\subsection{Analysis for Zero-Shot HOI Detection}
\begin{table}[t]
    \centering \small
    \caption{{Comparing the zero-shot performance with state-of-the-art methods on HICO-Det involves different settings.} In cases marked with *, it means that only detected boxes are used without object identification information from the detector. }\label{zeroshot_performance}
    \begin{tabular}{l>{\hfil}p{33pt}<{\hfil}>{\hfil}p{28pt}<{\hfil}>{\hfil}p{28pt}<{\hfil}>{\hfil}p{28pt}<{\hfil}}
\toprule
        Method & Type & Unseen & Seen & Full \\ \midrule
        Shen $et \  al.$ \cite{shen2018scaling} & UC & 5.62 & - & 6.26 \\ 
        FG \cite{BansalRSC20} & UC & 10.93 & 12.60 & 12.26 \\ 
        ConsNet \cite{LiuYC20} & UC & 16.99 & 20.51 & 19.81 \\ 
        \hline
        VCL \cite{hou2020visual} & RF-UC & 10.06 & 24.28 & 21.43 \\ 
        ATL \cite{HouY0PT21} & RF-UC & 9.18 & 24.67 & 21.57 \\ 
        FCL \cite{hou2021detecting} & RF-UC & 13.16 & 24.23 & 22.01 \\
        GEN-VLKT \cite{LiaoZLWL022} & RF-UC & 21.36 & 32.91 & 30.56 \\ 
        RLIPv2 \cite{yuan2023rlipv2} & RF-UC & 19.33 & 34.22 & 31.24 \\ 
        HOICIIP \cite{ning2023hoiclip} & RF-UC & 25.53 & 34.85 & 32.99 \\ 
        \rowcolor{blue!8} HOICIIP+ConCue  & RF-UC & \textbf{31.51} & \textbf{36.99} & \textbf{35.51} \\ 
        \hline
        VCL \cite{hou2020visual} & NF-UC & 16.22 & 18.52 & 18.06 \\ 
        ATL \cite{HouY0PT21} & NF-UC & 18.25 & 18.78 & 18.67 \\ 
        FCL \cite{hou2021detecting} & NF-UC & 18.66 & 19.55 & 19.37 \\ 
        GEN-VLKT \cite{LiaoZLWL022} & NF-UC & 25.05 & 23.38 & 23.71 \\ 
        RLIPv2 \cite{yuan2023rlipv2} & NF-UC & 21.18 & 28.95 & 27.40 \\
        HOICLIP \cite{ning2023hoiclip} & NF-UC & 26.39 & 28.10 & 27.75 \\ 
        \rowcolor{blue!8} HOICLIP+ConCue  & NF-UC & \textbf{30.54} & \textbf{32.45} & \textbf{32.07} \\ 
        \hline
        ATL* \cite{HouY0PT21} & UO & 5.05 & 14.69 & 13.08 \\ 
        FCL* \cite{hou2021detecting} & UO & 0.00 & 13.71 & 11.43 \\ 
        GEN-VLKT \cite{LiaoZLWL022} & UO & 10.51 & 28.92 & 25.63 \\ 
        HOICIIP \cite{ning2023hoiclip} & UO & 16.20 & 30.99 & 28.53 \\ 
        \rowcolor{blue!8} HOICLIP+ConCue  & UO & \textbf{18.64} & \textbf{33.78} & \textbf{31.18} \\ 
        \hline
        GEN\_VLKT \cite{LiaoZLWL022} & UV & 20.96 & 30.23 & 28.74 \\ 
        HOICIIP \cite{ning2023hoiclip} & UV & 24.30 & 32.19 & 31.09 \\ 
        \rowcolor{blue!8} HOICLIP+ConCue  & UV & \textbf{26.97} & \textbf{35.77} & \textbf{34.53} \\ 
\bottomrule
    \end{tabular}
\end{table}
By incorporating large VLMs, our approach benefits from the prior knowledge embedded within these models. Additionally, the contextual cues provide extra semantic information that can be used to infer relationships between objects and actions, making it more capable of handling unseen scenarios. 

Following~\cite{LiaoZLWL022, ning2023hoiclip}, we performed zero-shot experiments in three distinct evaluation settings: unseen combinations (UC), unseen verbs (UV), and unseen objects (UO). In the UC setting, the training data include all verb and object classes, but some HOI triplets remain unseen. We further divided this setting into two cases: Rare First UC (RF-UC), where the tail-end rare HOI classes are designated as unseen, and Non-rare First UC (NF-UC), where the head-end classes are designated as unseen. In both RF-UC and NF-UC settings, $120$ HOI classes are selected as unseen classes. In the UO setting, $12$ objects are randomly selected from a pool of $80$ object classes, and the HOI classes associated with these $12$ objects are treated as unseen classes during training. 
In the UV setting, $20$ verbs are randomly selected from a pool of $117$ verbs, and the HOI triples involving these $20$ verbs are treated as unseen during training.

From the experimental results shown in Table~\ref{zeroshot_performance}, HOICLIP outperforms other SOTA methods in various settings, demonstrating strong performance in handling unseen interaction categories. By integrating our ConCue approach, the HOICLIP+ConCue model significantly surpasses all baseline methods across multiple settings, including RF-UC, NF-UC, UO, and UV. 
Notably, in the RF-UC setting, HOICLIP+ConCue achieves a $5.98$ mAP improvement over the HOICLIP for the unseen classes. This remarkable improvement can be attributed to the incorporation of VLM within ConCue, which provides contextual cues that enhance interaction understanding. By leveraging the generalization capability of large VLMs, our method effectively boosts performance in these challenging settings.
In the NF-UC setting, where the challenge lies in large HOI combinations with sufficient samples set as unseen, the ConCue plugin still achieves remarkable performance improvements. For instance, performance on unseen categories improves by $15.7\%$ compared to HOICLIP, with a marginal improvement of $4.15$ mAP. This demonstrates the robustness of our method, even when dealing with complex and diverse unseen category combinations. Similarly,
in the UO and UV settings, where certain objects or verbs are designated as unseen, the ConCue plugin continues to significantly enhance HOICLIP's performance. These improvements reveal the effectiveness of using contextual cues, which facilitate model transfer and expansion to new, unseen categories. 
Similarly, ConCue consistently achieves higher mAP values for seen categories across all settings. This is particularly notable in the NF-UC setting, where it achieves a remarkable mAP gain of $4.35$. This substantial improvement further validates the effectiveness of our proposed approach. 

\subsection{Comparison with Large VLM}
In this paper, we utilize InstructBlip, a large VLM, to generate contextual visual cues for HOI detection. InstructBlip is a pre-trained model that learns joint image and text representations from large-scale data. Its notable performance has been demonstrated in various vision and language tasks, including image classification and image caption generation. 
The effectiveness of InstructBlip stems from its ability to identify and emphasize contextual cues within images, which is a central component of our method. To thoroughly assess the impact of our ConCue approach, it is essential to conduct a comparative analysis with the InstructBlip method.


\subsubsection{Prompt for InstructBlip} 
To guide InstructBlip in detecting the HOI triplet $<human, object, interaction>$ within an image, we devise a specific prompt. The details are as follows:

\begin{tcolorbox}[
    colframe=black!65,    
    colback=white,        
    coltitle=gray!50,       
    fonttitle=\bfseries,  
    title=HOI detection Prompt, 
    boxrule=0.5mm,        
    arc=1mm,              
    left=1mm, right=1mm, top=1mm, bottom=1mm, 
    width=\columnwidth,   
    enlarge left by=0mm,  
    ]

\textbf{Prompt:} Please analyze the image below and identify the interactions between person(s) and object(s) observed within the image.

\vspace{0em}

\textbf{$<$Output Format$>$}\\
Each response should be in the format: The person is [interaction label] [object label].

\vspace{0em}

\textbf{$<$Output Example$>$}
\begin{itemize}
    \item The person is playing a sports ball.
    \item The person is holding a laptop.
\end{itemize}

\end{tcolorbox}


\subsubsection{Performance Analysis} 
\begin{table}[t]
    \centering \small
        \caption{{Comparisons with InstructBlip and HOICLIP under the regular setting on HICO-Det.}}\label{blip1}
    \begin{tabular}{p{72pt}ccc}
    \toprule
        Method & Full & Rare & Non-Rare \\ \midrule
        InstructBlip & 29.67 & 15.47 & 30.21 \\ 
        HOICLIP & 72.37 & 18.24 & 74.37 \\ 
        \rowcolor{blue!8} 
        HOICLIP+ConCue & \textbf{75.82} & \textbf{29.31} & \textbf{77.58} \\ 
        \bottomrule
    \end{tabular}
\end{table}
The HOI detection task typically involves two subtasks: instance detection and interaction recognition. Although existing large VLMs can analyze images by managing interactions between visual content and textual prompts, they are not specifically tailored for the traditional HOI detection task. To compare InstructBlip with our ConCue, we omit the instance detection subtask and focus primarily on interaction recognition. Specifically, given an image $I$, our goal is to predict the interaction class between human-object pairs by directly generating the triplet $<human, object, interaction>$ from InstructBlip.

Table~\ref{blip1} presents the performance comparison between HOICLIP+ConCue, HOICLIP, and InstructBlip in the regular setting. It can be observed that the performance of HOICLIP and HOICLIP+ConCue outperforms InstructBlip across all three evaluation settings. This can be attributed to two key factors: (1) ConCue effectively integrates visual features with contextual cues derived from large VLMs for enhanced feature extraction, and (2) InstructBlip, as a general pre-trained model, is not specifically tailored for HOI tasks. While it can provide contextual understanding within an image, it struggles to accurately identify interactions between humans and objects.
Overall, the experimental results clearly demonstrate that HOICLIP+ConCue surpasses InstructBlip by a substantial margin, highlighting its effectiveness in incorporating contextual cues into conventional HOI detection methods. 
\subsection{Visualization}
\begin{figure*}[t]
\centering
\begin{minipage}{1.0\linewidth}\centering
\centerline{\includegraphics[height=11.0cm]{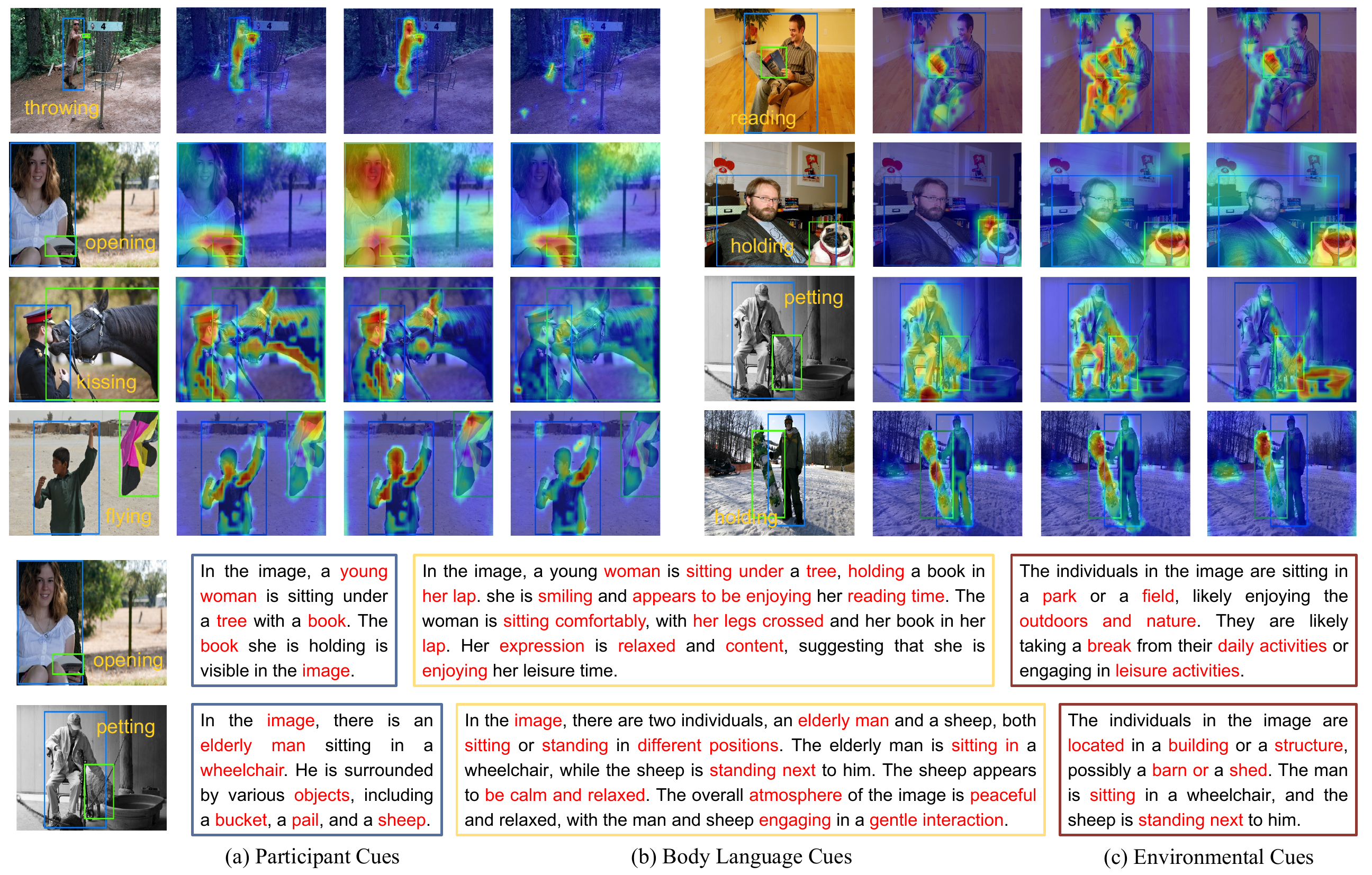}}
\end{minipage}
\vspace{0.5cm} 
\caption{Attention Visualization. The upper section shows the visualization of spatial feature attention in the interaction decoder, while the lower section displays the visualization of contextual visual cue attention in the decoder. In the upper section, the four images from left to right represent the prediction results, and the spatial attention maps for participant cues, body language cues, and environmental cues, respectively. In the lower section, words are highlighted in red if their attention exceeds the threshold.}\vspace{-0.0cm} 
\label{visualization_all}\medskip
\end{figure*}
We present the prediction results along with attention visualizations in Figure~\ref{visualization_all}. 
The upper part presents examples of spatial attention mAPs within the interaction decoder. The four images from left to right display predicted results, spatial attention mAPs for participants, body language, and environmental cues, respectively. It is evident that the ConCue plugin yields accurate predictions. When participant cues are provided, attention is more focused on identifying individuals and objects involved in human interactions. With body language cues, attention intensifies on facial features and hands. When environmental cues are introduced, attention disperses across the background of the image. The interaction decoder, guided by different cues, focuses on different regions. This highlights the necessity of not sharing parameters among the different decoders. Besides, it is easier for us to understand the rationale on which the model makes judgments regarding interactions, thereby enhancing the interpretability of the model.

Furthermore, we conducted an attention analysis of words in each cue for three of the images. Words exceeding the threshold are highlighted in red.
For instance, in the first image, ConCue extracts information from the participant's cues, indicating there is a woman, a book, and a tree in the image. In the body language cues generated by VLM, ConCue focuses on the person sitting under the tree, appearing relaxed and enjoying leisure time. Based on the environmental cues, the ConCue obtains the information that this is an outdoor park. Finally, ConCue gathers information from these cues and combines it with visual features to infer that the interaction type is ``a person \textcolor{red}{opening} a book''.

\subsection{Ablation Study}

\begin{table}[t]
    \centering \small
    \caption{The design of network architecture. The effectiveness of components is demonstrated by constructing each design on a $base$ model.}\label{ab}
    \begin{tabular}{p{109pt}ccc}
    \toprule
        Method & Full & Rare & Non-Rare \\ \midrule
        $base$ & 32.11 & 27.21 & 33.51 \\ 
        $+CLIP$ & 34.11 & 30.05 & 35.31 \\ 
        $+one\ tower$ & 35.46 & 33.57 & 35.97 \\
        $+mutil\ tower$ & 36.23 & 34.33 & 36.86 \\
        \rowcolor{blue!8} $+visual \ feature$ (ConCue) & \textbf{37.46} & \textbf{36.75} & \textbf{37.68} \\ 
        \bottomrule
    \end{tabular}
\end{table}
To highlight the effectiveness of our approach, we conducted ablation experiments with HOICLIP as the base model. We specifically examined the effects of the network architecture of ConCue, as well as the types and quantities of contextual cues.
\subsubsection{Effect of Network Architecture}
Initially, we developed a base model based on the DETR transformer architecture \cite{CarionMSUKZ20} and added an interaction decoder \cite{LiaoZLWL022}. The first variant, denoted as $+CLIP$, replaces the interaction decoder's input features with visual features from the CLIP encoder. Additionally, this variant employs the CLIP text encoder, incorporating prior knowledge to initialize the HOI classifier. This adaptation significantly enhanced performance, resulting in a $2.0$ mAP increase in the FULL setting.
The second variant, labeled $+one\ tower$, suggests that multiple contextual cues share a single tower structure. As shown in Table \ref{ab}, the $+one\ tower$ variant with contextual cues performs significantly better than $+CLIP$, indicating that contextual cues enhance interaction understanding. However, it does not perform as well as $+multi\ tower$.
The $+multi\ tower$ variant employs a multi-tower transformer architecture, assigning a dedicated transformer structure to each type of contextual cue. It shows a $0.77$ mAP increase in FULL setting compared with $+one\ tower$, this suggests that not sharing parameters between towers allows each tower to focus on different aspects of linguistic information, resulting in more robust decoder outputs.
Finally, we integrated visual features with contextual cue-aware visual features. In the FULL setting, this resulted in a $1.23$ mAP performance improvement. 

\begin{table}[t]
    \centering \small
        \caption{The design of contextual visual cues. The effectiveness of cues is validated from various perspectives. C1, C2, C3, and C4 represent participant cues, body language cues, environmental cues, and spatial cues, respectively.}\label{cue} 
    \begin{tabular}{>{\hfil}p{12pt}<{\hfil}>{\hfil}p{12pt}<{\hfil}>{\hfil}p{12pt}<{\hfil}>{\hfil}p{12pt}<{\hfil}>{\hfil}p{32pt}<{\hfil}>{\hfil}p{32pt}<{\hfil}>{\hfil}p{37pt}<{\hfil}}
    
\toprule
        C1 & C2 & C3 & C4 & Full & Rare & Non-Rare \\ 
        \midrule
        - & - & - & - & 33.79 & 29.65 & 34.78 \\ 
        \checkmark & - & - & - & 35.60 & 35.15 & 35.73 \\ 
        \checkmark & \checkmark & -  & - & 36.34 & 35.47 & 36.69 \\ 
        \checkmark & \checkmark & \checkmark & - & 36.98 & 36.02 & 37.32 \\
        \rowcolor{blue!8} \checkmark & \checkmark & \checkmark & \checkmark & \textbf{37.46} & \textbf{36.75} & \textbf{37.68} \\ 
        \bottomrule
    \end{tabular}
\vspace{-0.0cm}
\end{table}

\subsubsection{Effect of Contextual Cue Types and Quantities}
As shown in Table~\ref{cue}, we examined the significance of different types and quantities of cues. Specifically, C1, C2, C3, and C4 represent participant cues, body language cues, environmental cues, and spatial cues, respectively. All experiments were conducted under regular settings.
The results indicate that using information from a single cue leads to an approximate $5.50$ mAP gain in rare scenarios compared to methods that do not utilize contextual cues. This suggests that cues derived from large VLMs can assist the model in identifying interaction categories with insufficient data. 
When two cues are employed, the model's performance improves further, showcasing the additive benefit of incorporating multiple sources of information. However, the performance gain with two cues does not exceed that achieved with three cues, highlighting the unique contributions of each cue.
Integrating various cues enables our approach to leverage the strengths of each perspective, leading to a more comprehensive and robust interaction understanding. This also highlights the potential for further improvements through the strategic combination of multiple cues.


\subsection{Model Efficiency Analysis}
We conducted a comprehensive analysis of the training time for our model. The training process comprises two main phases: generating contextual cues using InstructBLIP and model training of ConCue and baselines.
The process of extracting multiple types of contextual cues using InstructBLIP took approximately 19 hours and was conducted on 4 A800 GPUs, ensuring efficient computation. This step provides a solid foundation and rich information for subsequent model training. 
Additionally, our ConCue plugin benefits from having fewer trainable parameters due to parameter freezing in the CLIP and RoBERTa components. As a result, when combined with baselines like HOICLIP and GEN-VLKT, our model shows significant efficiency, with the additional training time on the HICO-Det dataset being only one-sixth of the original model's training time.

%% file: sec/6_Conclusion.tex
\section{Conclusions and Future work}
In this paper, we propose an enhanced feature extraction approach for Human-Object Interaction (HOI) detection by leveraging the prior knowledge embedded in large VLMs. To achieve this, we introduce a novel method named ConCue, which integrates contextual information into instance and interaction decoders, enabling seamless integration with conventional HOI detection techniques.
Our approach employs a specialized prompt set designed to work with large VLMs to generate rich contextual cues from images, encompassing participant cues, body language cues, environmental cues, and temporal cues. To effectively utilize these contextual cues, we develop a transformer-based feature extraction module with a multi-tower architecture, which guides and enhances the feature extraction process.
Experimental results on two widely-used datasets demonstrate the superior performance of ConCue, showcasing its effectiveness in advancing the state-of-the-art in HOI detection.

While ConCue has made strides in HOI detection, several opportunities for future work remain. One direction is to combine conventional models with large Vision-Language Models (VLMs) to harness the strengths of both. Another focus is improving the quality of contextual cues by refining prompts and enhancing VLMs' contextual understanding. Additionally, ConCue could be adapted to related tasks like caption generation, broadening its applications in visual understanding.